\documentclass{article}
\usepackage[final]{nips_2018}
\usepackage[normalem]{ulem}
\usepackage{authblk}
\usepackage[utf8]{inputenc} 
\usepackage[T1]{fontenc}    
\usepackage{hyperref}       
\usepackage{url}            
\usepackage{booktabs}       
\usepackage{amsfonts}       
\usepackage{nicefrac,amsmath}       
\usepackage{microtype}      
\usepackage{subfigure}
\usepackage{array}
\usepackage{graphicx}
\usepackage{color}
\usepackage{siunitx}
\usepackage{textcomp}
\usepackage{gensymb}
\usepackage[square,sort,comma,numbers]{natbib}
\usepackage[font=scriptsize]{caption}

\title{Automating Motion Correction in Multishot MRI Using Generative Adversarial Networks}

\author[1,2]{Siddique Latif}
\author[1]{Muhammad Asim}
\author[1]{Muhammad Usman}
\author[1]{Junaid Qadir}
\author[2]{Rajib Rana}

\affil[1]{Information Technology University (ITU)-Punjab, Pakistan}
\affil[2]{University of Southern Queensland, Australia}

\begin{document}

 \maketitle
\begin{abstract}
Multishot Magnetic Resonance Imaging (MRI) has recently gained popularity as it accelerates the MRI data acquisition process without compromising the quality of final MR image. However, it suffers from motion artifacts caused by patient movements which may lead to misdiagnosis. Modern state-of-the-art motion correction techniques are able to counter small degree motion, however, their adoption is hindered by their time complexity. 
This paper proposes a Generative Adversarial Network (GAN) for reconstructing motion free high-fidelity images while reducing the image reconstruction time by an impressive two orders of magnitude. 
\end{abstract}
\section{Introduction}


Multishot Magnetic Resonance Imaging (MRI) has been an excellent imaging modality which provides high resolution image while addressing some well-known problems of classical standard MRI techniques---such as the slowness of true single-echo sequences and the problems of inflexible contrast and limited resolution in single-shot sequences \cite{bernstein2004handbook}. However, it suffers from a critical drawback: it is highly sensitive towards subjects' motion; even very small movement (e.g., due to the respiration or slight head motion) can become a limiting factor for ultra-high resolution images. 
The common presence of motion artifacts in Multishot MR image---which results from the dislocation of the \textit{k}-space that is filled fractionally in each shot---is clearly undesirable especially in clinical settings due to the risk of misdiagnosis \cite{andre2015toward} and must be addressed. Although modern state-of-the-art motion correction techniques (e.g., the iterative algorithms proposed in \cite{cordero2016sensitivity,loktyushin2015blind}) are able to counter smaller degree of motion, their adoption is hindered by the inordinate time for motion correction. 
We address this void by proposing an end-to-end trained Generative Adversarial Network (GAN) for motion correction together with a popular parallel MRI reconstruction algorithm known as Conjugate Gradient Sensitivity Encoding (CG SENSE) to return superior results. \textit{Our results show that by removing the motion estimation step involved in the current schemes, and by employing an end-to-end trained GAN, motion artifacts can be faithfully removed to produce high quality images while reducing the image construction time.}


\begin{figure}[!ht]
\centering
\centerline{\includegraphics[width=0.8\textwidth]{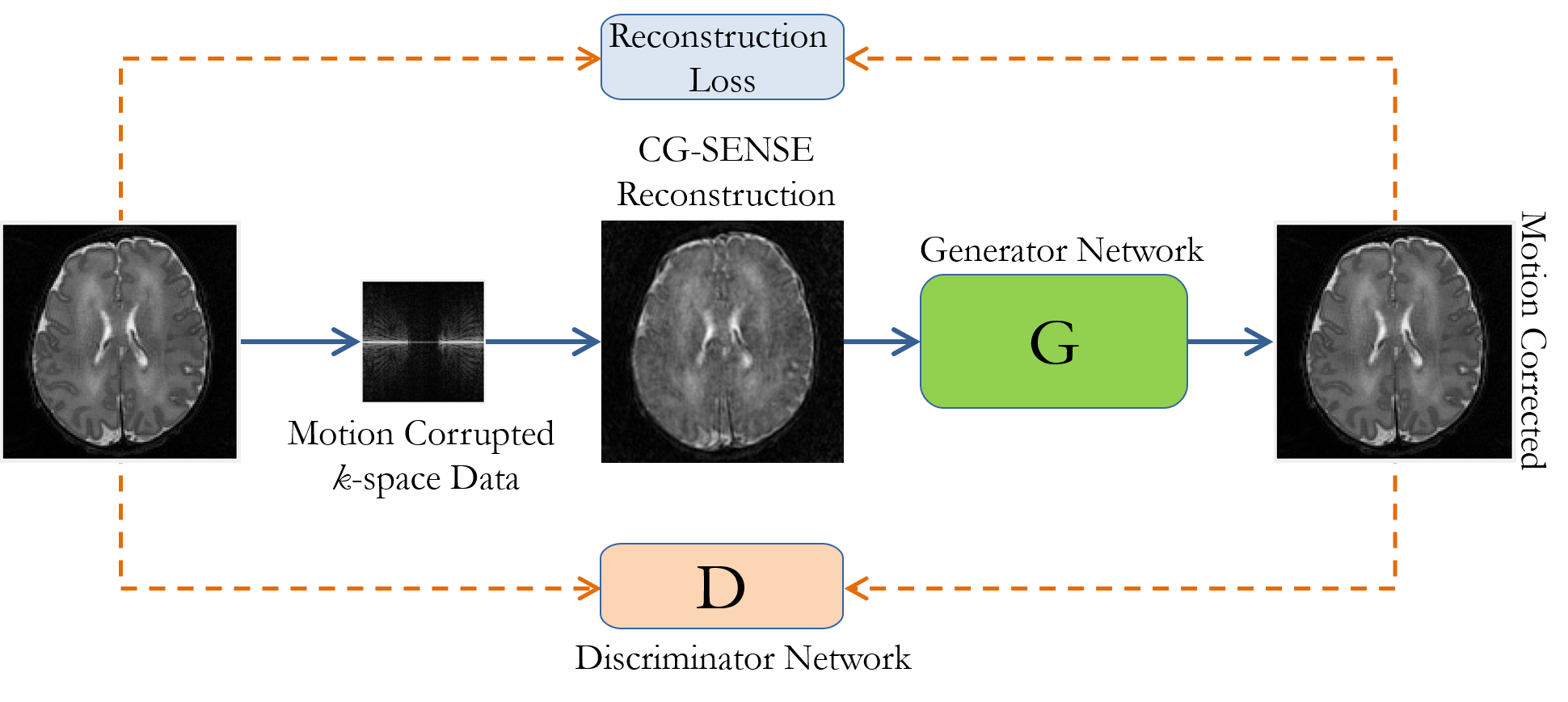}}
\captionsetup{width=0.95\textwidth}
\caption{The proposed motion correction framework for multishot MRI, where CG SENSE is 
used to reconstruct the motion-corrupted images, and the generator network of the GAN, 
in conjunction with the discriminator network, is tasked with motion correction.} 
\label{fig:proposed}
\end{figure}

\section{Model Architecture and Dataset}

\begin{figure}
    \centering
    \includegraphics[width=.8\textwidth]{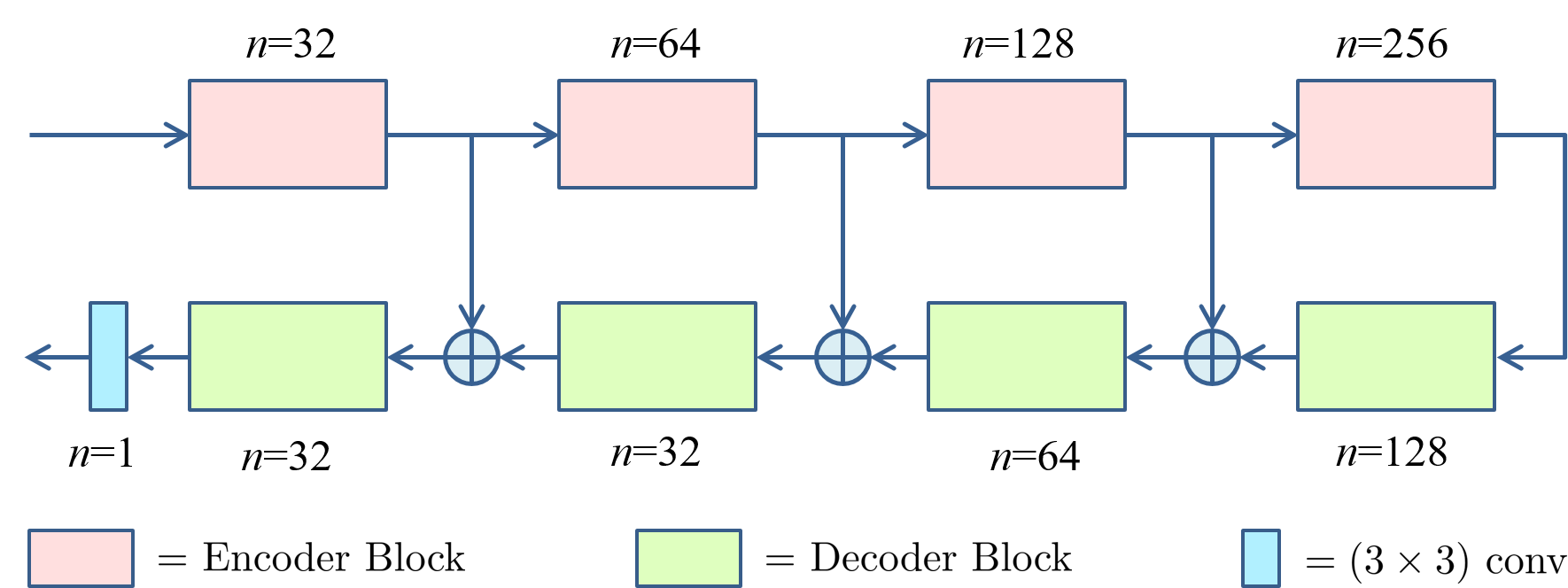}
    \captionsetup{width=.8\linewidth}
    \caption{Architecture of the generator network (U-Net) consisting of encoding and decoding blocks. The discriminator has exactly the same architecture as the encoder part of our U-Net.}
    \label{fig:model_arch}
\end{figure}

Our choice of a GAN-based model for motion correction stems from their great success in image restoration problems and even in MRI reconstruction \cite{mardanisemi}, where they have shown their capabilities in producing high-fidelity images \cite{pan2018physics}. The GAN framework includes a generator $G$ and a discriminator $D$ which play an adversarial game defined by the following optimization program. 
\begin{equation}
    \underset{G}{\text{min}} \  \underset{D}{\text{max}} \quad \mathrm{E}_x[\log(D(x))] + \mathrm{E}_y[\log(1 - D(G(y)))]
\end{equation}
where $y$ and $x$ denote the motion corrupted and corrected images, respectively. 
We implemented U-Net \cite{ronneberger2015u} like 
convolutional network architecture for the generator consisting of an encoder and decoder as shown in Figure \ref{fig:model_arch}. 
Every encoder block consists of $5$ convolutional layers (each with $n$ feature maps) except for the middle layer that has $n/2$ feature maps. 
The decoder blocks have similar structure as the encoder blocks except that we replace all the convolutional layers with transposed convolutions.

 Since there is no public dataset available for motion-corrupted images, we rely on synthetically generated data for evaluating the performance of our proposed method. We have utilized images from the BRATS 2015 dataset \cite{menze:hal-00935640}. 
 Motion artifacts have been introduced in motion-free images of the dataset by employing the technique proposed in \cite{bydder2002detection}, and also utilized by 
 \cite{cordero2016sensitivity} and \cite{loktyushin2015blind} to evaluate the performance of motion correction techniques. This method consists of three phases: \textit{firstly}, the introduction of motion in motion-free \textit{k}-space data for each shot; \textit{secondly}, the extraction of specific segments of \textit{k}-space data; and \textit{lastly}, the combination of segmented \textit{k}-space data. Since the BRATS dataset \cite{menze:hal-00935640} is in the image domain, therefore, the additional step of the creation of simulated \textit{k}-space multishot MRI data, following \cite{allison2013accelerated}, has been performed. 
 
 \begin{figure}
    \centering
    \raisebox{0.3in}{\rotatebox[origin=t]{90}{Corrupted}}
    \subfigure{\includegraphics[width=0.18\textwidth]{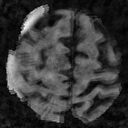}}
    \subfigure{\includegraphics[width=0.18\textwidth]{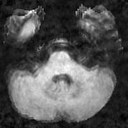}}
    \subfigure{\includegraphics[width=0.18\textwidth]{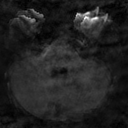}}
    \subfigure{\includegraphics[width=0.18\textwidth]{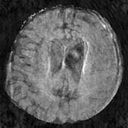}}
    \subfigure{\includegraphics[width=0.18\textwidth]{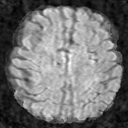}}\\[-0.9em]
    \raisebox{0.3in}{\rotatebox[origin=t]{90}{Cordero et al.}}\hspace{0.08em}
    \subfigure{\includegraphics[width=0.18\textwidth]{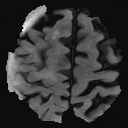}}
    \subfigure{\includegraphics[width=0.18\textwidth]{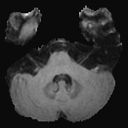}}
    \subfigure{\includegraphics[width=0.18\textwidth]{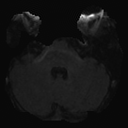}}
    \subfigure{\includegraphics[width=0.18\textwidth]{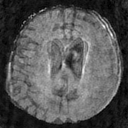}}
    \subfigure{\includegraphics[width=0.18\textwidth]{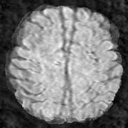}}\\[-0.9em]
    \raisebox{0.3in}{\rotatebox[origin=t]{90}{Ours}}\hspace{0.08em}
    \subfigure{\includegraphics[width=0.18\textwidth]{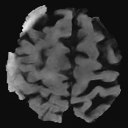}}
    \subfigure{\includegraphics[width=0.18\textwidth]{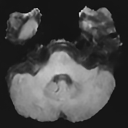}}
    \subfigure{\includegraphics[width=0.18\textwidth]{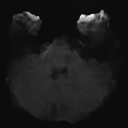}}
    \subfigure{\includegraphics[width=0.18\textwidth]{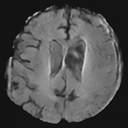}}
    \subfigure{\includegraphics[width=0.18\textwidth]{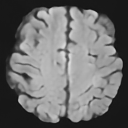}}\\[-0.9em]
    \setcounter{subfigure}{0}
    \raisebox{0.30in}{\rotatebox[origin=t]{90}{Original}}
    \subfigure[\ang{5}]{\includegraphics[width=0.18\textwidth]{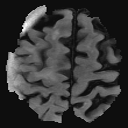}}
    \subfigure[\ang{8}]{\includegraphics[width=0.18\textwidth]{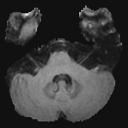}}
    \subfigure[\ang{10}]{\includegraphics[width=0.18\textwidth]{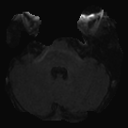}}
    \subfigure[\ang{12}]{\includegraphics[width=0.18\textwidth]{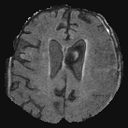}}
    \subfigure[\ang{14}]{\includegraphics[width=0.18\textwidth]{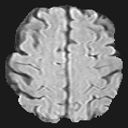}}
    
     \label{fig:2shot_mri}
    \caption{Qualitative results of our solution's output (third row) for increasing degrees of motion \ang{5}, \ang{8}, \ang{10}, \ang{12}, and \ang{14} compared with on motion-corrupted MRI images (first row), groundtruth images (bottom row), and state-of-the-art algorithm \cite{cordero2016sensitivity}'s output (second row). \textit{It can be seen that the output of our algorithm is qualitatively better than the output of previous approach.}}
\end{figure}  

\section{Experiments and Discussion}
The simulated \textit{k}-space motion-corrupted data has been reconstructed using standard CG SENSE algorithm \cite{pruessmann1999sense}.  
For each dataset, corresponding to a certain degree of motion, we train our network using RMSProp optimizer with learning rate $1 \times 10^{-4}$ and batch size of $16$, until convergence. For each update of $G$, we updated $D$ twice. In each case, we also pre-train the generator network $G$ to allow training to converge faster. 
The network was trained on $70\%$ of data (13019 images) and all results are reported using remaining 30\% of test data (5580 images).
\begin{table}[ht]
\centering
\scriptsize
\begin{tabular}{|l|l|l|l|l|l|}
\hline
Degree of motion & \ang{5} & \ang{8} & \ang{10} & \ang{12} & \ang{14} \\ \hline
Peak Signal to Noise Ratio (PSNR) & 31.82 & 28.27 & 27.64 & 27.20 & 26.82 \\ \hline
Structural Similarity Index (SSIM) & 0.95 & 0.92 & 0.91 & 0.90 & 0.90 \\ \hline
\end{tabular}
\vspace{.5mm}
\caption{Average values of performance metrics using GAN for 2 shot MRI imaging on entire test data.}
\label{table:psnr-ssim-ap}
\end{table}

For the smaller degree of motions, our network successfully captures the underlying statistical properties to recover sharp and excellent images (see Figure 3). Also, high average values for Peak Signal to Noise Ratio (PSNR) and Structural Similarity Index (SSIM) on the entire test data are observed at \ang{5} in Table \ref{table:psnr-ssim-ap}. We observe a smooth performance decay with increasing degrees of motion, as expected, and at \ang{14}, MRI scans are corrupted to an extent where they would be rendered practically useless. Our network removes most of the corruptions, but due to severe degradation, recovers overly smooth images. We choose to compare our approach with the current state of the art method \cite{cordero2016sensitivity}. Notably, this method was able to recover from up to \ang{10} rotations in 2-shot acquisitions, and we are achieving significantly acceptable quality of MR images up to \ang{14} using GANs.  

\begin{table}[!ht]
\centering
\scriptsize
\captionsetup{width=0.9\textwidth}
\begin{tabular}{|m{2.8cm}|m{1cm}|m{1cm}|m{1cm}|m{1cm}|m{1cm}|}
\hline
Degree of Motion
& \ang{5} & \ang{8} & \ang{10} & \ang{12} & \ang{14} 
\\ \hline
\begin{tabular}[c]{@{}l@{}}Cordero et al. \cite{cordero2016sensitivity} \end{tabular}
&\begin{tabular}[c]{@{}l@{}}25.23 \end{tabular}
&\begin{tabular}[c]{@{}l@{}}39.27\end{tabular}
&\begin{tabular}[c]{@{}l@{}}57.93\end{tabular}
&\begin{tabular}[c]{@{}l@{}}134.27\end{tabular}
&\begin{tabular}[c]{@{}l@{}}152.03\end{tabular}
\\\hline
\begin{tabular}[c]{@{}l@{}}Our Approach \end{tabular}
&\begin{tabular}[c]{@{}l@{}}0.22\end{tabular}
&\begin{tabular}[c]{@{}l@{}}0.24\end{tabular}
&\begin{tabular}[c]{@{}l@{}}0.24\end{tabular}
&\begin{tabular}[c]{@{}l@{}}0.25\end{tabular}
&\begin{tabular}[c]{@{}l@{}}0.34\end{tabular}
\\\hline
\end{tabular}
\vspace{.5mm}
\caption{Comparing computational time in seconds of our approach with the current state-of-art technique \cite{cordero2016sensitivity}. The reported times include both reconstruction (using CG SENSE) as well as motion correction time (using GAN) on the entire test data.}
\label{table:time}
\end{table}
We see through our results documented in Table \ref{table:time} that we are correcting \ang{5} motion at almost $114$ time faster then the  Cordero et al. \cite{cordero2016sensitivity} method 
due to the outsourcing of the motion-correction task (which was previously estimated iteratively during with the significant delay) to the GAN. 

\section{Conclusions}
The work in this paper is the first instance of employing a GAN for motion correction in MRI.
It works on top of the standard CG SENSE reconstruction similar to the previous method on motion correction \cite{cordero2016sensitivity}. However, it does not estimate motion states prior to its correction, which significantly lessens the computational time that is very crucial for clinical applications. We empirically showed with our preliminary results that generative framework provided by GANs can provide significantly improved results in terms of time complexity and clinically acceptable image quality. 
In this work we have reported results for $2$ shot acquisition, we are aiming in our future work to apply GANs for an increased number of shots and to perform a more extensive set of experiments. 







\end{document}